\documentclass[preprint,3p,twocolumn]{elsarticle}

\usepackage{amssymb}
\usepackage{graphicx}
\usepackage{amsfonts}
\usepackage[vlined,ruled,linesnumbered]{algorithm2e}
\usepackage{subcaption}
\usepackage{booktabs}
\usepackage{mathtools}
\usepackage{enumitem}
\usepackage[dvipsnames]{xcolor}

\DeclareMathOperator*{\argmin}{arg\,min}
\newcommand{\fig}[1]{Fig.~\ref{fig:#1}}

\begin{document}

\begin{frontmatter}



\title{Not All Datasets Are Born Equal: \\On Heterogeneous Data and Adversarial Examples}


\cortext[cor1]{Corresponding author}
\fntext[fn1]{Contributed equally.}

\affiliation[inst1]{organization={Department of Software and Information Systems Engineering, \\Ben-Gurion University of the Negev},
            city={Beer-Sheva},
            postcode={8410501}, 
            country={Israel}}
            
\author[inst1]{Yael Mathov\corref{cor1}\corref{cor2}\fnref{fn1}}\ead{yaelmath@post.bgu.ac.il}
\author[inst1]{Eden Levy\fnref{fn1}}\ead{leed@post.bgu.ac.il}
\author[inst1]{Ziv Katzir}\ead{zivka@post.bgu.ac.il}
\author[inst1]{Asaf Shabtai}\ead{shabtaia@bgu.ac.il}
\author[inst1]{Yuval Elovici}\ead{elovici@bgu.ac.il}

\begin{abstract}
Recent work on adversarial learning has focused mainly on neural networks and domains where those networks excel, such as computer vision, or audio processing.
The data in these domains is typically homogeneous, whereas heterogeneous tabular datasets domains remain underexplored despite their prevalence.
When searching for adversarial patterns within heterogeneous input spaces, an attacker must simultaneously preserve the complex domain-specific validity rules of the data, as well as the adversarial nature of the identified samples. 
As such, applying adversarial manipulations to heterogeneous datasets has proved to be a challenging task, and no generic attack method was suggested thus far. 
We, however, argue that machine learning models trained on heterogeneous tabular data are as susceptible to adversarial manipulations as those trained on continuous or homogeneous data such as images.
To support our claim, we introduce a generic optimization framework for identifying adversarial perturbations in heterogeneous input spaces.
We define distribution-aware constraints for preserving the consistency of the adversarial examples and incorporate them by embedding the heterogeneous input into a continuous latent space.
Due to the nature of the underlying datasets We focus on $\ell_0$ perturbations, and demonstrate their applicability in real life.
We demonstrate the effectiveness of our approach using three datasets from different content domains.
Our results demonstrate that despite the constraints imposed on input validity in heterogeneous datasets, machine learning models trained using such data are still equally susceptible to adversarial examples. 
\end{abstract}




\end{frontmatter}



\section{Introduction}
The susceptibility of learning algorithms to adversarial input manipulations has puzzled researchers in recent years, leading to extensive research work.
Early work on adversarial attacks was demonstrated using learning tasks that were popular at the time, such as spam filtering \citep{dalvi2004adversarial,lowd2005adversarial} and malware detection \citep{biggio2013evasion}. 
Notably, homogeneous, bag-of-words feature representation was used in all relevant use cases. 
Later on, as deep learning has become prevalent, the vast majority of papers published in recent years have discussed attacks and defenses in the context of neural networks.
Naturally, these works have focused on input domains where deep learning excels such as image and audio, where data representation remains mainly homogeneous and input spaces are continuous.
A significant amount of research effort was put into computer vision applications and related tasks, e.g., image segmentation \cite{arnab2018robustness}, face detection \cite{sharif2016accessorize}, and more specifically,
image classification \cite{szegedy2013intriguing,goodfellow2014explaining,papernot2016limitations,kurakin2016adversarial,carlini2017towards,su2019one}.
Other domains include audio \cite{carlini2018audio}, and the exclusively discrete domains of malware represented by binary vectors \cite{grosse2016adversarial,hu2017generating} and the classification of sparsely encoded text \cite{jia2017adversarial,ebrahimi2018hotflip}.

In contrast, there has been little to no research on \textit{heterogeneous} input spaces in the context of adversarial examples.
The reasons for this range from the intrinsic diversity of feature data types, to the relatively underexplored robustness of the tree-based models that are most associated with such data \cite{shavitt2018regularization,arik2019tabnet,popov2019neural}. 
The few attacks that were demonstrated in the context of heterogeneous input domains have focus on exploiting the architecture of the learning model, while ignoring the effects of the manipulation on the validity of the data itself.
Another possible explanation is that machine learning research in recent years was almost entirely focused on deep learning based methods, leading researchers to study data taken from high-dimensional, homogeneous and continuous distributions, where deep learning typically demonstrates improved performances over traditional approaches \cite{nielsen2015neural}.
Conversely, tabular data is drawn from low-dimensional, heterogeneous, and largely discrete input domains.
More specifically, tabular data has the following characteristics: (a) the input often includes a combination of nominal, ordinal, and real-valued features, (b) different value ranges are associated with different features, (c) missing values are common, (d) some features are considered immutable in the context of an attack, and (e) complex cross-feature interactions define valid and anomalous feature combinations.
Examples for domains where the data is normally heterogeneous include healthcare, real estate, and financial applications.

In this paper, we study the process of crafting adversarial examples for heterogeneous tabular data.
There are several challenges that are unique to this case, both from the perspective of the adversary and challenges that are associated with the optimization procedure of finding adversarial perturbations that must comply with the heterogeneity.
In contrast to audio or language processing applications that are considered digital, machine learning tasks on heterogeneous data are usually related to real-life scenarios, and the relevant features describe tangible information, such as a person's salary.
Therefore, an adversary that targets heterogeneous data would like to minimize the number of modified features (i.e., a minimal $\ell_0$ perturbation), as opposed to minimizing other frequently used distance metrics (e.g., $\ell_1, \ell_2, \ell_\infty$) which are compromised by the vastly different feature scales.
Additionally, some input features must be considered immutable in the context of the attack, as it is impossible (or at least very difficult) to change their values in real life. 
As an example consider features that denote a person's height, credit history, or residential address.  
This implies that an adversary must also address the real life \textit{feasibility} of the identified adversarial perturbations.
Data heterogeneity lead to optimization challenges as well. 
A generic attack algorithm must preserve the \textit{consistency} of the adversarial examples in terms of feature correlation, and to maintain the various types of features (e.g., integers, positives, real numbers).

Our approach for perturbing samples from heterogeneous input spaces addresses each of the aforementioned challenges, while being agnostic to the target learning algorithm and the data domain.
First, we formulate the mathematical modeling of the \textit{validity} of the data, in terms of its consistency (i.e., preserving feature correlation) and feasibility (i.e., keeping immutable features intact).
Then, we aim to capture the complexity of the heterogeneous input space by using a \textit{data transformation} function, whose purpose is to embed the heterogeneous data into a continuous latent space and lead to more consistent adversarial examples.
This function encompasses the correlations between the features and thus allows us to use gradient-base optimization techniques to identify adversarial examples without harming the consistency of the input.
Using these tools and techniques, we define a general optimization problem for finding valid adversarial examples of heterogeneous tabular data, with the ultimate goal of applying them in real life.
Finally, we propose a clear implementation of our approach and its various components, based on neural networks and methods from metric learning.
We test our approach and implementation using three datasets from different content domains and a variety of target learning algorithms.

Our results suggest that machine learning models in domains with heterogeneous input spaces are just as prone to adversarial manipulations as those used in continuous homogeneous domains. 
This implies that further research is required to understand the role played by data, as opposed to the model or algorithm, in the context of adversarial susceptibility.

\section{Background}
Tree-based models achieve state-of-the-art performance in learning tasks that involve heterogeneous tabular data \cite{shavitt2018regularization,popov2019neural,arik2019tabnet}.
The robustness of such models has been extensively researched, including attacks on decision stumps or trees \cite{papernot2016transferability,andriushchenko2019provably,chen2019robust}, random forests \cite{kantchelian2016evasion,chen2019robust}, and gradient boosting machines \cite{cheng2019query,chen2019robust}.
Similarly to these works, prior studies on traditional non-tree-based models were either too specific to the domains they were demonstrated in \cite{dalvi2004adversarial,kolcz2009feature,biggio2010multiple} or tailored to a certain class of learning algorithms (e.g., SVM or linear models in general) \cite{lowd2005adversarial,biggio2011support}.
The most prominent attack methods today target neural networks \cite{szegedy2013intriguing,goodfellow2014explaining,papernot2016limitations,kurakin2016adversarial,carlini2017towards}.
The vast majority of these attacks relate to a more general technique based on the network's gradient or its approximation.
However, in the case of nominal features which are common in tabular data, the derivative with respect to such features (i.e., $\frac{\partial\mathcal{L}}{\partial x_i}$ for a loss function $\mathcal{L}$ and a nominal feature $x_i$) cannot be correctly interpreted, as it is arbitrary at best.
As a result of this and other issues, all previously mentioned research either lacks generality or could not be adapted out-of-the-box to applications with heterogeneous input space.
We aim to bridge this gap by extending the renowned adversarial deep learning tools to create a neural network-based implementation of our generic approach.

Recently, there have been some attempts at manipulating neural networks trained on heterogeneous tabular data.
\citet{sarkar2018robust} performed a number of attacks on a credit score classifier and demonstrated vulnerabilities in financial applications.
However, the experiment only included continuous features despite the heterogeneous data domain, and the evaluation metrics are not well-suited to a real-life adversary.
\citet{ballet2019imperceptible} presented a real-life data-oriented approach that could be adapted to domains other than finance.
It employs the means to (1) preserve the data types, by clipping, and (2) reduce the perturbation's perceptibility, by incorporating feature importance in the attack objective in an attempt to mimic the decision process of a domain expert.
Their method and experiment did not appropriately reflect the heterogeneity of the data, since categorical features were treated as continuous, and nominal features were discarded altogether.
Moreover, the evaluation of the perturbations was based on the $\ell_2$ norm, which is misrepresentative given the significant variability in scales of the features, and the feature importance (i.e., knowledge of a domain expert) was improperly modeled by using the Pearson correlation with the target variable.

A primary goal of our work is to cover all aspects related to the feature space challenges of heterogeneous tabular data.
The following works integrate components related to covariances or outliers with the intent of preserving feature correlations or feature types.
An early work \cite{biggio2013evasion} in the adversarial learning domain takes the underlying distribution of the data into account in their proposed attack.
Their method includes modeling the a priori estimate (i.e., $p(x \,|\, y)$ for a data point $x$ and its label $y$) by using a kernel density estimator as an additional component in their attack objective.
However, this method was not evaluated on heterogeneous input spaces and therefore lacks additional related aspects that are essential.
\citet{ustun2019actionable} formulated an integer linear programming (ILP) problem to find perturbations that do not include immutable features and are consistent (i.e., preserving the data's structure under its various mathematical constraints) by adding constraints to the problem definition.
\citet{kanamoridace2020dace} extended this work by incorporating non-linear statistical elements that correspond to the empirical distribution of the data, in a mixed ILP-based optimization instance.
Their method aims to create realistic perturbations that conform with feature correlations and outlier values, by adding approximations of the Mahalanobis' distance \cite{mahalanobis1936generalized} and the local outlier factor \cite{breunig2000lof} to their objective function and optimization constraints.
The context of the two last-mentioned works is \textit{counterfactual explanation} methods \cite{wachter2018counterfactual} of machine learning models used to provide recourse to users who wish to gain access to the factors underlying the model's decision (e.g., a credit scoring application where customers try to improve their score).
Contrarily, our work focuses on an adversarial setting and related topics.

The approach we propose in this paper complements previous work by weaving the various concepts we discussed so far into one exhaustive formulation and implementation, and therefore the approach addresses all possible challenges that arise from crafting adversarial examples for heterogeneous data.
To the best of our knowledge, we are the first to introduce a comprehensive and general adversarial attack method that is agnostic to the target learning algorithm and focuses on truly heterogeneous tabular input space.

\section{Our Approach}
In this section, we describe our approach using the following notations.
The attacker targets a machine learning model $M:\mathcal{X} \rightarrow \mathcal{Y}$.
For brevity, we assume $\mathcal{X}$ is a discrete input space with $d$ discrete random variables.
The label space is denoted by $\mathcal{Y} \subseteq \{0, 1\}^m$ where labels are one-hot encoded.

Let $X_1,\dots,X_d$ be the random variables that represent the features in the data, which have the discrete distributions $\mathcal{F}_1,\dots,\mathcal{F}_d$, such that for each, $1\leq i\leq d$, $X_i\sim\mathcal{F}_i$.
Therefore, an input vector $X=(X_1,\dots,X_d)$ follows the multivariate joint distribution $\mathcal{F}$, as in $X\sim\mathcal{F}$.
We note that $X_1,\dots,X_d$ are not necessarily independent or identically distributed.
It would be even safer to assume some level of dependency between the features as naturally prevails in tabular data.

Given a benign sample $x \in \mathcal{X}$, with the corresponding ground true label $y \in \mathcal{Y}$, a maximal perturbation size $\lambda$, and a context dependent distance metric $\mathcal{D}$, the adversarial goal is to find an adversarial example $x + \delta = x^* \in \mathcal{X}$ such that  $\mathcal{D} \left (x^*, x \right ) < \lambda$ and $M(x^\ast) \neq y$.

We assume that the attacker has access to the training data $\mathbb{X} \subseteq \mathcal{X}$ used for constructing $M$.
Although our approach introduces an untargeted attack, it can be easily modified to fit a targeted attack.

\subsection{Definitions for Data Validity}\label{sec:app-integrity}
To ensure the validity of $x^*$, the attack algorithm must preserve the distributional consistency of the various features and also refrain from modifying immutable features so that the resulting perturbation pattern remains feasible in real life.
These two challenges are of vital importance and enable the attacker to eventually apply the adversarial perturbations in practice (i.e., in the real world).

\subsubsection{Data Consistency}
Changing a sample $x$ with the corresponding label $y$ might result in values that are unlikely given $\mathcal{F}$; i.e., values that lie in unsupported regions of $\mathcal{F}$.
In such a case, we refer to $x^*$ as inconsistent and therefore consider it invalid.
Formally, for a consistency threshold $\epsilon > 0$, we consider a sample to be \textit{$\epsilon$-inconsistent} if the probability $\mathbb{P}_{X\sim\mathcal{F}} (X=x^* \,|\, y) \leq \epsilon$.
Note that consistency is measured in the context of a specific class label, so an input vector can be highly likely given one label but be anomalous for another.

Two examples of inconsistency are described as follows:
\begin{enumerate}
    \item A scenario in which $x_i^* \in x^*$ where $x_i^*$ is an undefined outcome of $X_i^*$ or $\mathbb{P}_{X_i\sim\mathcal{F}_i} (X_i=x_i^* \,|\, y)=0$; for instance, $X_i$ represents the number of children but $x_i^* \notin \mathbb{N}$.
    \item For each $x_i^* \in x^*$, we have $\mathbb{P}_{X_i\sim\mathcal{F}_i} (X_i=x_i^* \,|\, y)>0$ but $\mathbb{P}_{X\sim\mathcal{F}} (X=x^* \,|\, y) \leq \epsilon$; consider $X_i, X_j$ to be year of birth and year of death respectively but $x_i^* > x_j^*$.
\end{enumerate}

Consequently, we define $x^*$ as an $\epsilon$-consistent adversarial example if the following holds:
\begin{equation}\label{eq:consistent}
    \mathbb{P}_{X\sim\mathcal{F}} (X=x^* \,|\, y)> \epsilon
\end{equation}

\subsubsection{Data Feasibility}
Data describing real-life use cases involves features that represent somewhat tangible details.
In real-world applications, some of the information collected is factual and immutable, meaning it cannot be changed by an attacker.
Although it is possible to digitally perturb immutable features, they cannot be changed in real life and any change to them would render the example unfeasible and therefore invalid.
Examples for immutable features include a patient's blood type, a customer's history of financial transactions, and the size of a piece of real estate.

We define the subset of indexes of immutable features as $I \subseteq \{1, \dots ,d \}$, and require an adversarial example $x^*$ to satisfy the following, in order to be feasible:
\begin{equation}\label{eq:immutable}
    \forall i \in I,\; x_i^*=x_i
\end{equation}

\subsubsection{Data Validity}
A valid adversarial example $x^*$ must satisfy both Equations~\ref{eq:consistent} and~\ref{eq:immutable} to be consistent and feasible. 
Therefore, for a consistency threshold $\epsilon$ and set of immutable features $I$, the validity of an adversarial example $x^*$ is modeled as the conjunction:
\begin{equation}\label{eq:valid}
    \mathbb{P}_{X\sim\mathcal{F}} (X=x^* \,|\, y)> \epsilon \; \land \; \forall i \in I,\;x_i^* = x_i
\end{equation}

\subsection{Data Transformation}\label{sec:datarestructuring}
Unlike continuous data, categorical and nominal features impair the performance of commonly used optimization techniques, such as the training of neural networks \cite{nielsen2015neural}.
As a result, many methods for crafting adversarial examples are limited when used in conjunction with heterogeneous tabular data.
In addition, perturbing a non-continuous feature without taking into account the support of its corresponding distribution is highly likely to result in an illegal value.
This puts the attack method in jeopardy and may result in $\epsilon$-inconsistent and illegal adversarial examples.

To address these challenges, we suggest using a structure-preserving transformation function, which practically serves as an embedding function.
It is defined as $f: \mathcal{X} \rightarrow \mathbb{R}^e$.
This new representation provides two advantages: 
\begin{enumerate}
    \item The process of constructing $f$ induces the learning of latent patterns of features that are otherwise less adequately represented; i.e., categorical and nominal features.
    Such patterns include interactions among features, semantic meanings of nominal features, and more generally, the modeling of the probability distribution of the data.
    The parameters of $f$ carry meaningful information and allow the gradients of the network to encompass these patterns, which in turn, assists the optimization process in finding consistent adversarial examples.
    \item The data is transformed into a continuous and homogeneous space, making it easier to manipulate using optimization-based attacks and measure distances.
\end{enumerate}

The embedding function $f$ can be implemented using various general concepts, including generative adversarial networks \cite{goodfellow2014generative}, autoencoders \cite{rumelhart1985learning}, and metric learning \cite{weinberger2009distance}.

\subsection{Attack Description}\label{sec:approach-attack}
Recall that the attacker's goal is to craft adversarial examples for a target model $M: \mathcal{X} \rightarrow \mathcal{Y}$.
The target model is trained on the heterogeneous dataset $\mathbb{X} \subseteq \mathcal{X}$, which represent real-world information.
Therefore, $M$ could be based on any machine learning algorithm, however it is more likely that $M$ is tree-based, as this is the type that is most associated with heterogeneous tabular data.
Additionally, perturbing a benign input sample $x \in \mathcal{X}$ should craft a valid adversarial example $x^* \in \mathcal{X}$; hence, the attack preserves the validity of the data.

\begin{figure*}
    \centering
    \includegraphics[width=0.9\textwidth]{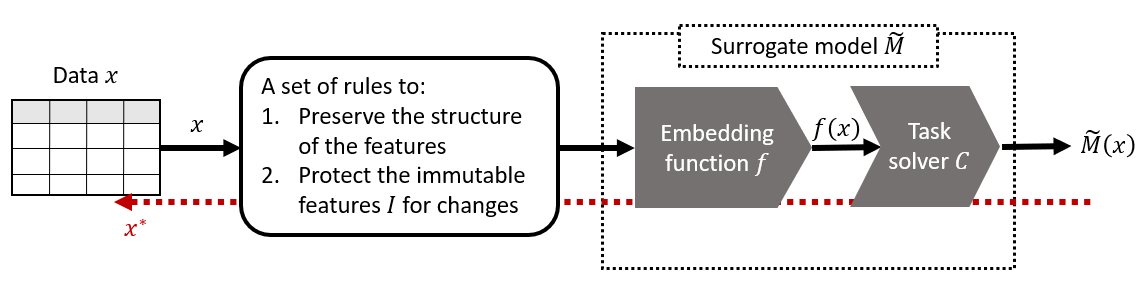}%

    \caption{The suggested attack approach:
    A set of rules are applied to the tabular data $x$, which is sent into an embedding function $f$.
    The embedded data $f(x)$ is sent to a learning-based task solver $C$, which outputs a prediction similar to the target model $M$.
    Since our design uses $f$ to transform $x$, $M$ cannot be used as a task solver; so instead, we use the surrogate model $\widetilde{M} = C \circ f$ (the dotted rectangle).
    When using a gradient-based method to craft adversarial examples (the dashed line), the rules and the embedding function preserve the validity of the perturbed data.
    }
    \label{fig:attack}
\end{figure*}

Therefore, the proposed attack method (presented in \fig{attack}) uses two components that are designed to preserve the validity of the data: an embedding function to conserve the consistency of the data, and a set of rules that protect the immutable features (i.e., create a feasible $x^*$).
The embedding function is the structure preserving transformation $f: \mathcal{X} \rightarrow \mathbb{R}^e$.
Since it transform each input sample $x \in \mathcal{X}$ into a new continuous space, $M$ cannot be used to classify $f(x)$.
We address this challenge by using the knowledge on $\mathbb{X}$ to build a surrogate model $\widetilde{M}: \mathcal{X} \rightarrow \mathcal{Y}$ that mimics the predictions of $M$.
The architecture of $\widetilde{M}$ includes the embedding function and a classifier that receives the embedded data samples as an input. 
Then, we use $\widetilde{M}$ to craft the adversarial examples.

Since we craft adversarial examples for $\widetilde{M}$ and not for $M$, our attack algorithm relies on the transferability property \cite{papernot2016transferability} of adversarial examples.
The attacker crafts adversarial examples for a surrogate model $\widetilde{M}$ and uses them against the target model $M$.
The surrogate is created to support the requirements of the data transformation process and to facilitate the generation of adversarial examples that comply with the constraints set forth in Equation~\ref{eq:valid}.

\subsubsection{Constructing a Surrogate Model}
To preserve the consistency of the data, a surrogate model $\widetilde{M}: \mathcal{X} \rightarrow \mathcal{Y}$ is constructed by composing a classifier $C$ and an embedding function $f$, such that $\widetilde{M} = C \circ f$.
The structure preserving function $f$ transforms the data, and $C: \mathbb{R}^e \rightarrow \mathcal{Y}$ uses the embedded input samples to solve the original learning task.  
It is important to note that $f$ and $C$ are trained separately and sequentially, with $C$ training over $\{f(x) \;|\; x \in \mathbb{X} \}$.
Similarly to the embedding function $f$, the task solver can be implemented using multiple techniques.
However, such implementation should comply with requirements of the attack method and the structure of the embedded data.

\subsubsection{Generating Adversarial Examples}
Let $x \in \mathcal{X}$ be a valid input sample, such that $y \in \mathcal{Y}$ is the true label of $x$.
Given a consistency threshold $\epsilon > 0$, a set of immutable features $I \subseteq \{1, \dots ,d \}$, a distance metric $\mathcal{D}$, and a maximum perturbation size $\lambda$, the attacker aims to find a valid adversarial example $x^\ast \in \mathcal{X}$, such that $M(x^*) \approx \widetilde{M}(x^*) \neq y$.
Due to the design of $\widetilde{M}$ and its components, which are differentiable by our definition, the attacker can also use any optimization-based attack to craft adversarial examples and transfer them to the target model.

Formally, based on Equation~\ref{eq:valid}, finding a valid $x^*$ is done by solving the following optimization problem:
\begin{equation}
    \label{eq:attack-opt}
    \centering
    \begin{tabular}{p{0.1\textwidth}p{0.1\textwidth}}
        \multicolumn{2}{c}{$\argmin\limits_{x^*}\ \ \mathcal{L}_{adv}(\widetilde{M}(x^*),\,y) + \|f(x)-f(x^*)\|_2$}\\[0.25cm]
            \multicolumn{1}{r}{\text{s.t. }}& \multicolumn{1}{l}{$\mathbb{P}_{X\sim\mathcal{F}} (X=x^*\,|\,y) > \epsilon,$} \\[0.15cm]
            &\multicolumn{1}{l}{$\forall i \in I,\; x_i^* = x_i,$} \\[0.15cm]
            &\multicolumn{1}{l}{$\mathcal{D}(x,\,x^*) < \lambda$}
    \end{tabular}
\end{equation}
where $\mathcal{L}_{adv}$ is defined as the adversarial objective function.
The attacker chooses $\mathcal{L}_{adv}$ according to his goals, capabilities, and data domain.
For example, $\mathcal{L}_{adv}$ can combine the original task's loss function (e.g., cross-entropy, mean squared error) with additional domain specific objectives, such as the real-life complexity in changing the values of individual input features.
The second component in the objective, defined as $\mathcal{L}_{spatial}=\|f(x)-f(x^*)\|_2$, helps arrive at a closer point in the latent space, which results in more consistent adversarial examples by the construction of $f$.

The optimization problem in Equation~\ref{eq:attack-opt} is not a simple task, especially because of its validity constraints.
Calculating $\mathbb{P} (X=x^*\,|\,y)$ cannot be done trivially, but it can be approximated by using CTGAN~\cite{xu2019modeling}, which is specific to heterogeneous tabular data; or other more general approaches, such as Bayesian networks~\cite{zhang2017privbayes} and decision trees~\cite{reiter2005using}.
As $\epsilon$ increases, the adversarial example that is $\epsilon$-consistent is more likely and therefore more imperceptible in real life.
However, there is a clear trade-off between finding such an $\epsilon$-consistent adversarial example and its likelihood, as this optimization problem is highly non-convex, and its convergence is not guaranteed.
The feasibility constraint can be forced by applying a feature mask that prevents perturbations to immutable features from being added.

These challenges require a more eased optimization instance that approximates the original problem and can be solved by using existing algorithms.
We propose an implementation of such an optimization instance in our experiments which are described in the next section where we begin by describing the learning process of a surrogate model, which includes an embedding function as outlined in Section \ref{sec:datarestructuring}, and a task solver.
Then we illustrate the complete course of the adversarial attack algorithm, as well as the means employed to enforce the data validity rules.

\section{Experiments}
\subsection{Datasets}
We conducted our empirical study on three tabular datasets from different content domains.
These datasets were chosen because they all have the various challenging characteristics of tabular data, and especially because they are heterogeneous and pertain to real-life oriented tasks.

\begin{itemize}
    \item Home Credit Default Risk (HCDR) \cite{hcdr} - includes current and historical financial information on just over $300,000$ customers and their loan requests. This dataset's use case is a binary classification task in the credit risk assessment area designed to predict a client's repayment abilities.
    \item Intensive Care Unit (ICU) \cite{icu} - contains medical data on around $91,000$ patients in an intensive care unit. This dataset is associated with a binary classification task aimed at helping predict the death of ICU patients.
    \item U.S. Lodging Listings (Airbnb) \cite{lairbnb} - data scrapped from the Airbnb website on some $57,000$ property listings in major U.S. cities. This dataset's use case is a regression task aimed at predicting the price of lodging.
\end{itemize}

In the binary classification tasks, both of the datasets suffer from class imbalance with a ratio of approximately $92:8$.
The global minimum and maximum values in each dataset can be extremely far apart, with the difference between them ranging from approximately $10^3$ in Airbnb to $10^5$ in ICU and $10^9$ in HCDR.
Missing values are quite common as well, with ICU having nearly a quarter of all of the data missing, down to $15\%$ in HCDR and $9\%$ in Airbnb.
HCDR, ICU, and Airbnb include $22$, $81$, and $48$ immutable features, respectively.
Table~\ref{tbl:ds-table} lists the number of constraints imposed on the features in each dataset by constraint type and the total number of features.

\begin{table*}[t]
  \centering
  \begin{tabular}{lrrrrrr}
    \toprule                 
    Dataset     & Integer & Positive & Negative & Normalized & Categorical & Total Features\\
    \midrule
    HCDR    & 11    & 11    & 5     & 15    & 21    & 52 \\
    ICU     & 42    & 104   & 0     & 1     & 21    & 126 \\
    Airbnb  & 20    & 22    & 1     & 5     & 133   & 157 \\
    \bottomrule
  \end{tabular}
  \caption{Number of constraints by type (features may have more than one constraint).}
  \label{tbl:ds-table}
\end{table*}

As commonly done for these data domains, we applied the following preprocessing steps on the datasets.
First, the data was cleaned of outliers and features for which over 75\% of the values were missing.
In cases in which there was a group of features with unusually high correlation of over 0.95 or under -0.95, we dropped all but one randomly chosen feature from the group.
Missing values in each of the remaining features were imputed with one of the following: the modal value for categorical variables, the mean value for regular numeric variables, or a unique value outside the variable's legitimate value range.
We chose the combination of methods that yielded the best performance.
Categorical features were encoded with consecutive integers, so that all of the data is numeric.
Finally, all of the data was re-scaled or standardized so that each feature is in the range $[0, 1]$ or normally distributed.

Note that our preprocessing maintains the challenging characteristics of tabular datasets, and the attack method backpropagates through the relevant operations.
Each learning algorithm has its own capabilities, limitations, and implementation; therefore, some of the algorithms do not require all of the above preprocessing.
For example, the LightGBM algorithm~\cite{ke2017lightgbm} handles missing values and categorical variables as per its implementation, so we skipped the relevant preprocessing steps for that model.

\subsection{Attack Implementation}\label{sec:exp-impl}
The following is an implementation of the various components of our approach.
It is important to note that other choices can be made for any component and part of the process as long as the general requirements are met, including the models, data transformation, validity rules, method for perturbing the input samples, etc.

\subsubsection{Models}\label{sec:exp-models}
Here we describe the target and surrogate models used in the experiments.
All of the datasets were split into training and validation sets consisting of 80\% and 20\% of the data, respectively.
Another separate set of 500 samples was used to craft the adversarial examples.

\textbf{Target models.}
As previously noted, heterogeneous tabular data is more commonly associated with tree-based models, such as decision tree and random forest, as well as gradient boosting machines like LightGBM.
Due to their popularity, we demonstrate the transferability of our attack on the surrogate model to these three learning algorithms.
The performance evaluation of the models for each dataset are presented in Table~\ref{tbl:target-models-eval}.

\begin{table*}[t]
  \centering
  \begin{tabular}{llrrrrrrrr}
    \toprule
    \multicolumn{1}{c}{}&
    \multicolumn{1}{c}{}&
    \multicolumn{2}{c}{Decision Tree} &
    \multicolumn{2}{c}{Random Forest} &
    \multicolumn{2}{c}{GBM} &
    \multicolumn{2}{c}{Surrogate} \\
    \cmidrule(r){3-4}
    \cmidrule(r){5-6}
    \cmidrule(r){7-8}
    \cmidrule(r){9-10}
         Dataset & Metric & Train & Validation & Train & Validation & Train & Validation & Train & Validation\\
    \midrule
    HCDR & AUC    & 0.72  & 0.69 & 0.74  & 0.72 & 0.78  & 0.75 & 0.74  & 0.73    \\
    ICU & AUC     & 0.83 & 0.83 & 0.87  & 0.87 & 0.95  & 0.91 & 0.90  & 0.88      \\
    \midrule
    Airbnb & MSE  & 0.13 & 0.14 & 0.13  & 0.13 & 0.07  & 0.08 & 0.11  & 0.11  \\
    \bottomrule
  \end{tabular}
  \caption{Evaluation scores of the training and validation sets on the target and surrogate models for each task. The performance is evaluated using the ROC-AUC metric for classification tasks and the mean squared error (MSE) for regression.}
  \label{tbl:target-models-eval}
\end{table*}

\textbf{Surrogate model.}
The surrogate model consists of an embedding network and a task solver network.
Motivated by concepts from the person re-identification task, in each embedding network we use batch hard triplet loss \cite{weinberger2009distance,hermans2017defense} with the cosine similarity distance metric.
To fit the regression task to the same embedding learning process used in classification, we use equal frequency binning on the target variable.
We train the embedding networks with the Adagrad optimizer~\cite{duchi2011adaptive}, a learning rate of 0.001, and a mini-batch size of 32.
In our experiments, the size of the embedding vectors for HCDR, ICU, and Airbnb is four, 20, and 15, respectively.
Then, we freeze the embedding network's parameters and add the task solver sequentially; it consists of a single layer with one neuron and a sigmoid or softmax activation function.
The solver receives the embedding vectors from the trained embedding network as input.
We use the Adam optimizer~\cite{kingma2014adam} to minimize the standard objectives matching the learning task: cross-entropy for classification and MSE for regression.

\subsubsection{Attack Overview}\label{sec:exp-attack}
Our attack implementation, presented in Algorithm~\ref{alg:adv}, is a variant of the $\ell_0$-oriented Jacobian saliency map attack~\cite{papernot2016limitations}.
This iterative algorithm performs the following main steps.
First, it selects the feature that has the most impact on the prediction of the model, excluding immutable features.
To avoid having the algorithm repetitively make changes that will be clipped at the end of the iteration, features that have already been selected are not reconsidered.
The impact should be measured according to the model that is being attacked; in the case of our neural network-based surrogate model, it is the gradients, as described in Algorithm~\ref{alg:adv}, and for the tree-based target models it is the information gain of each feature.
Second, a perturbation vector $\alpha$ for the selected features is calculated by using the Adam optimizer which minimizes the adversarial objective function, as in Equation~\ref{eq:attack-opt}.
The auxiliary method for this calculation is denoted by \textsc{ComputeStep}.
Lastly, the \textsc{Project} method is performed on $x^*$, projecting each feature value onto its consistent set or continuous range, and ensuring that for each feature $i$, $\mathbb{P}_{X_i\sim\mathcal{F}_i} (X_i=x_i^*)>0$ (though not ensuring that $\mathbb{P}_{X\sim\mathcal{F}} (X=x^*)>0$).
We set the maximum perturbation size $\lambda$ to be the number of \textit{mutable} features.
The first part in the exit condition of the algorithm is different for regression tasks: $|\widetilde{M}(x^*) - y| > \tau$, for a predefined threshold $\tau$.
Given that the target variable in the Airbnb use case is the logarithm of the property's price and considering its variance, we set $\tau = 0.75$.

\SetKwInput{KwInput}{Input}
\begin{algorithm*}[t]
    \DontPrintSemicolon
    \KwInput{$\widetilde{M}$, $x$, $y$, $\mathcal{L}$, $\mathcal{L}^*$, $I$, $\lambda$}
     $x^* \leftarrow x$\;
     $S \leftarrow \emptyset$\;
     
     \While{$\widetilde{M}(x^*) = y$ \textbf{and} $|S| < \lambda$}{
      $G \leftarrow \nabla _{x^*} \mathcal{L} (\widetilde{M}(x^*), y)$\;
      $S \leftarrow S \cup \{\arg \max_{i \notin S \cup I} G_i\}$\;
      
      $\alpha \leftarrow \textsc{ComputeStep}(x^*, \mathcal{L}^*)$\;

      \ForAll{$i \in S$}{
      $x_i^* \leftarrow x_i^* + \alpha_i$\;
      }
      
      $x^* \leftarrow \textsc{Project}(x^*)$\;
     }
     \textbf{return} $x^*$
     \caption{\textbf{Crafting an adversarial example.} $\widetilde{M}$ is the surrogate model, $x$ is the candidate benign sample, $y$ its label, $\mathcal{L}$ is the original task's loss function, $\mathcal{L}^*$ is the adversarial objective function composed of $\mathcal{L}_{adv}$ and $\mathcal{L}_{spatial}$, $I$ is the set of immutable features, and $\lambda$ is the maximum distortion ($\ell_0$-wise).}
    \label{alg:adv}
\end{algorithm*}

\section{Results}
The surrogate and target models are evaluated using the standard metrics relevant to the task: the receiver operating characteristic-area under the curve (ROC - AUC) for imbalanced classification and the MSE for regression.
The performance of the models presented in Table~\ref{tbl:target-models-eval} shows that the GBM model consistently outperforms every other model in every use case and that its performance is also very close to highly-ranked submissions in the corresponding Kaggle competitions (apart from the Airbnb which is not part of a Kaggle competition).
It is important to note that our surrogate model does not fall too far behind the GBM model, which means that the embedding sub-network did not affect the overall performance too negatively.

\begin{figure*}
    \centering
    \subfloat[HCDR]{%
        \includegraphics[width=0.33\textwidth]{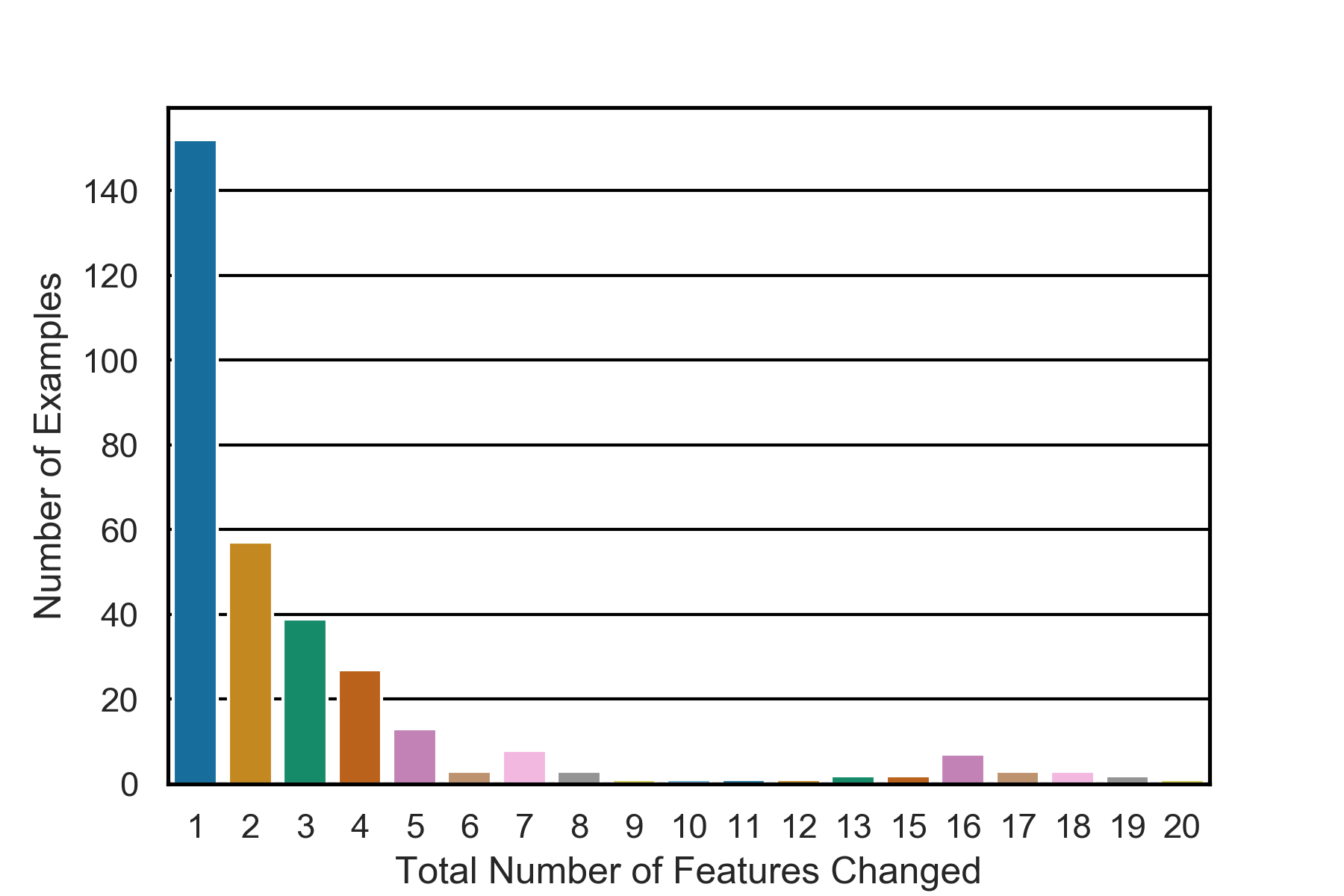}%
        }%
    \hfill%
    \subfloat[ICU]{%
        \includegraphics[width=0.33\textwidth]{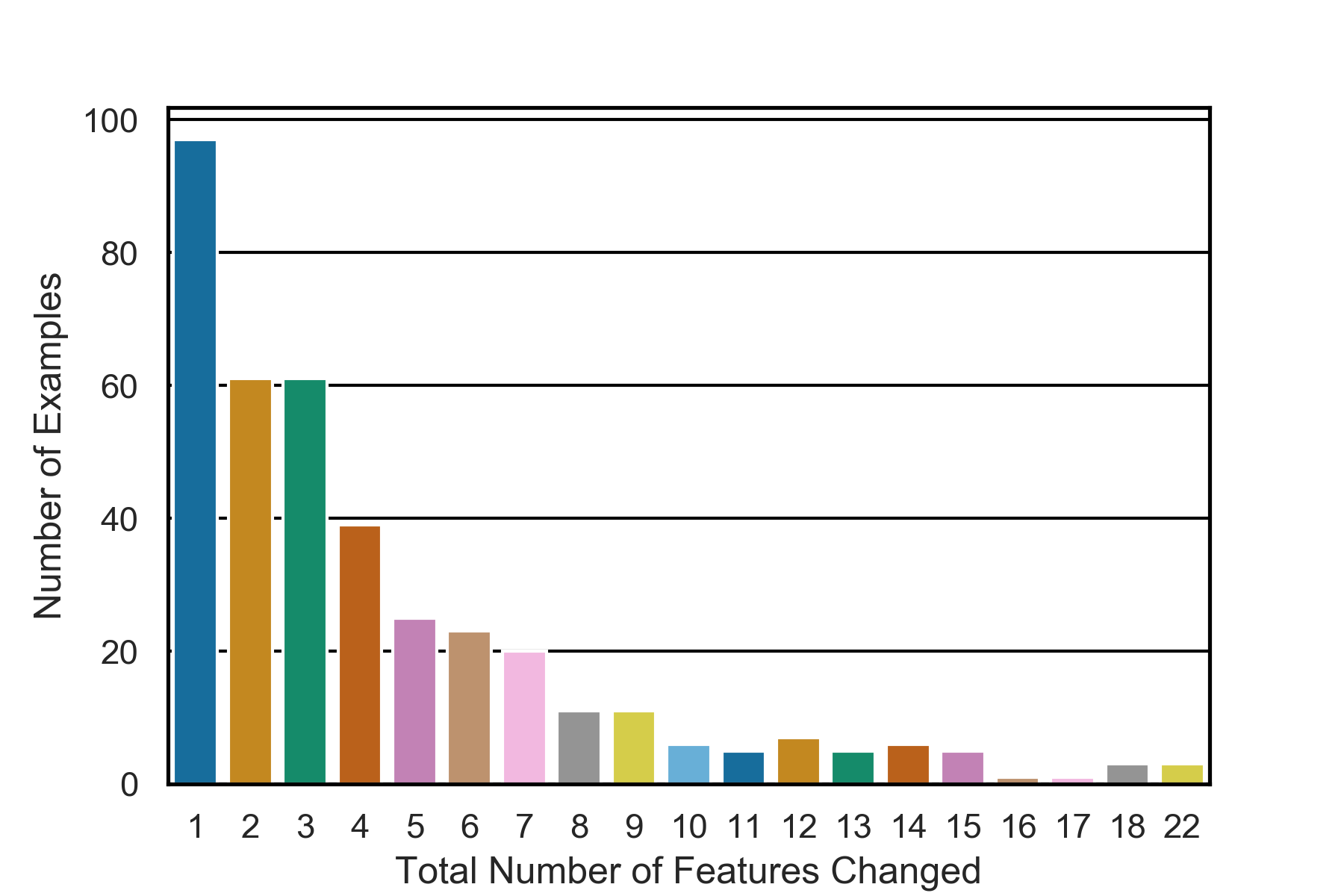}%
        }%
    \hfill%
    \subfloat[Airbnb]{%
        \includegraphics[width=0.33\textwidth]{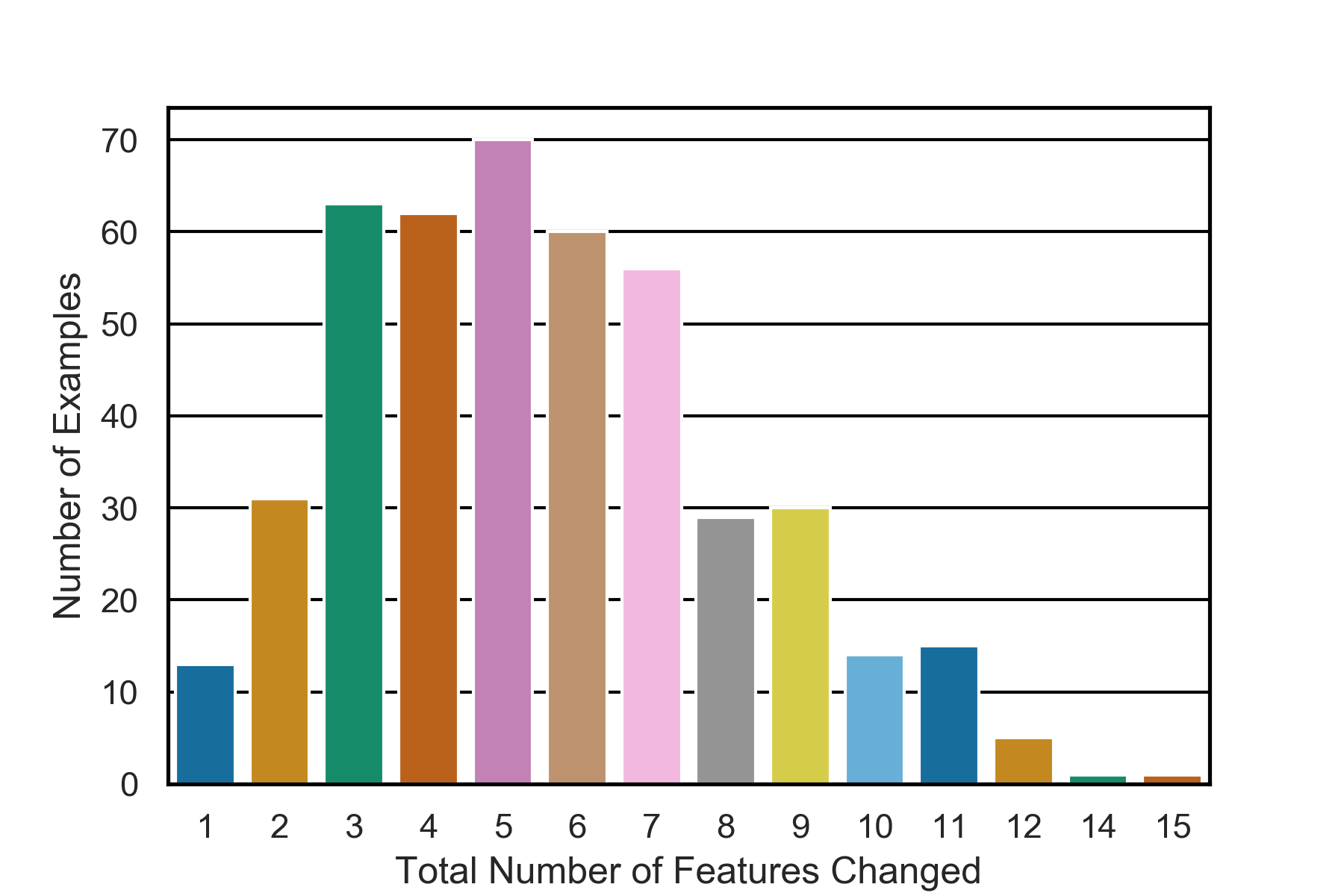}%
        }%
    \caption{Number of adversarial examples by the total number of features changed (i.e., $\ell_0$ distance).}
    \label{fig:l0hist}
\end{figure*}

As we have established, the metric most relevant to the case of heterogeneous data and real-world use cases is the $\ell_0$ norm, as the other norms are compromised by the significantly different scales of features.
However, we also decided to include the $\ell_1$ norm in our analysis for just numeric features in the perturbations, in order to obtain some understanding of how these features are perturbed in cases in which there are unusually big changes.
In addition to the $\ell_0$ norm, we also looked at its percentage of the entire dataset's features, as the number of features varies between the use cases.
An example of the need to analyze the percentage can be found in the Airbnb use case where the value of the average total $\ell_0$ distance is $5.28$, but when incorporating the total number of features in the dataset, which is $157$, we get an average \textit{relative} $\ell_0$ (i.e., total $\ell_0 (\%)$) of only $3.36\%$.
In Table~\ref{tbl:adv-analysis} we can see that in every use case the bulk of the average perturbation, are changes in categorical features.
Moreover, since the Airbnb dataset contains considerably more categorical, which are mostly binary features (representing the existence of amenities in the property), the difference is even more noticeable, with $4.33$ of the total $\ell_0$ distance of $5.28$ being changes in categorical features.
\fig{l0hist} presents histograms of the $\ell_0$ norms of the perturbations, which are highly right-skewed and indicate that the vast majority of the adversarial examples include changes to only 1-2 features.
The success rate is consistently high throughout the use cases, with nearly all of the samples in the attack set becoming adversarial examples.


\begin{table*}[t]
  \centering
  \begin{tabular}{lrrrrrrrr}
    \toprule
    \multicolumn{1}{c}{} &
    \multicolumn{2}{c}{Categorical} &
    \multicolumn{3}{c}{Numeric} &
    \multicolumn{2}{c}{Total}&\\
    \cmidrule(r){2-3}
    \cmidrule(r){4-6}
    \cmidrule(r){7-8}
         Dataset & $\ell_0$ & $\ell_0$ (\%) & $\ell_1$ & $\ell_0$ & $\ell_0$ (\%) & $\ell_0$ & $\ell_0$ (\%) & Success\\
    \midrule
    HCDR    & 1.60 & 7.62\% & 6.71 & 0.67 & 2.16\% & 2.27 & 4.37\% & 98.20\%   \\
    ICU     & 1.80 & 8.56\% & 4.96 & 1.64 & 1.56\% & 3.51 & 2.79\% & 94.40\%   \\ 
    Airbnb  & 4.33 & 3.26\% & 4.25 & 0.95 & 3.96\% & 5.28 & 3.36\% & 99.80\%   \\
    \bottomrule
  \end{tabular}
    \caption{
  Analysis of the adversarial attack on the surrogate model.
  The metrics correspond to the average across the perturbations of successful adversarial examples.
  The $\ell_0$ value is partitioned into categorical and numeric features.
  The $\ell_0$ (\%) column relates to the corresponding part of the $\ell_0$ from the relevant type of features.
  The success rate is relative to the attack set consisting of $500$ samples in each task.}
  \label{tbl:adv-analysis}
\end{table*}

In addition to attacking the surrogate model, we examine the transferability of the adversarial examples to GBM, random forest, and decision tree models.
As described in Algorithm~\ref{alg:adv}, the iterative feature selection is based on the gradients of the surrogate network.
When targeting models other than the surrogate, this selection process can be adjusted so that the impact of each feature is represented explicitly by the model, rather than implicitly by a surrogate's gradients.
We changed it to be the feature importance of the tree-based models we target, namely the information gain of each feature.
In Table~\ref{tbl:transferability} we report the transferability rates of the adversarial examples for each of the target models, as well as the rates of the adjusted attack (i.e., when feature selection is based on the target model's feature importance) in parentheses.
The base transferability rates in all use cases and models is already high, however the adjustment made evident and substantial improvements consistently throughout our experiments.

\begin{table*}[t]
  \centering
  \begin{tabular}{lrrr}
    \toprule
    Dataset &
    \multicolumn{1}{c}{GBM} &
    \multicolumn{1}{c}{Random Forest} &
    \multicolumn{1}{c}{Decision Tree} \\
    \midrule 
    HCDR    & 38.29\% (44.26\%)   & 35.85\% (42.22\%)   & 38.20\% (47.50\%)\\
    ICU     & 27.00\% (36.32\%)   & 20.40\% (36.32\%)   & 23.60\% (60.53\%)\\
    Airbnb  & 14.42\% (29.78\%)   & 18.83\% (32.93\%)   & 21.36\% (31.50\%)\\
    \bottomrule
  \end{tabular}
  \caption{Ratios of successfully transferred adversarial examples that were generated in an attack on the surrogate model. The parentheses indicate the transferability rates of attacks which were adjusted to each target model.}
  \label{tbl:transferability}
\end{table*}

In our experiments, the specific compositions of the various datasets (i.e., the size of groups of feature types) and the distribution of the constraints (as described in Table~\ref{tbl:ds-table}) have a direct effect on the adversarial optimization process.
Generally, we observed that heterogeneous input spaces involve several constraints and pose many challenges to the adversarial susceptibility of learning algorithms.
Our results show that not only is it possible to overcome such challenges and attack neural networks trained on heterogeneous tabular data very successfully and effectively, but that the transferability property of the adversarial examples generated is rather persistent in this unique case as well.

\section{Conclusions}
In this paper, we explore the role played by data in the susceptibility of machine learning models to adversarial manipulations.
We introduce a general approach that formalizes the generation of valid adversarial examples for \textit{heterogeneous} tabular data, and provide an implementation of such an attack in an optimization-based context.
Our approach is focused on an $\ell_0$-oriented adversary and her ability to apply the adversarial perturbations in real life.
To the best of our knowledge, this is the first study to present an automated attack for heterogeneous tabular data.
Our results support the possibility of successfully attacking previously unexplored domains with heterogeneous input space.

Future work may place greater emphasis on the important aspect of preserving feature correlation and improving our implementation, as well as developing ways to properly evaluate the success of such preservation.
Additional implementations for our approach should also be explored, mainly different embedding functions and other methods for crafting adversarial examples.
Advancement of mitigation methods could have more potential in the case of heterogeneous data than in the general case, as the baseline naive approach would be to limit the training to immutable features.

We believe that our work has identified an important research direction in the field of adversarial learning and broadens its scope beyond the main applications of computer vision.

\bibliographystyle{elsarticle-num-names}
\bibliography{cas-refs}





\end{document}